\documentclass{article}

\PassOptionsToPackage{square,sort,comma,numbers}{natbib}

\usepackage[preprint]{nips_2018}

\usepackage{hyperref}
\usepackage{url}

\usepackage[utf8]{inputenc} 
\usepackage[T1]{fontenc}    
\usepackage{booktabs}       
\usepackage{amsfonts}       
\usepackage{nicefrac}       
\usepackage{microtype}      

\usepackage{amsfonts}
\usepackage{hyperref}
\usepackage{graphicx}
\usepackage{url}
\usepackage{bm}
\usepackage[export]{adjustbox}
\usepackage{amsmath}
\usepackage{amsbsy}
\usepackage{calc}
\usepackage{color}

\usepackage{subcaption}

 \newcommand{\jcom}[1]{\textcolor{orange}{}}
 \newcommand{\maicom}[1]{\textcolor{red}{}}
 \newcommand{\fcom}[1]{\textcolor{green}{}}
 \newcommand{\scom}[1]{\textcolor{green}{}}

\newcommand{\de}{d(E)}

\newtheorem{theorem}{Theorem}[section]

\title{The Relationship Between High-Dimensional Geometry and Adversarial Examples}

\author{Justin Gilmer, Luke Metz, Fartash Faghri, Samuel S. Schoenholz, Maithra Raghu,\\ \textbf{Martin Wattenberg, \& Ian Goodfellow} \\
Google Brain\\ \texttt{\{gilmer,lmetz,schsam,maithra,wattenberg,goodfellow\}@google.com} \\ \texttt{faghri@cs.toronto.edu}}

\begin{document}

\maketitle

\begin{abstract}
    Machine learning models with very low test error have been shown to be consistently vulnerable to small, adversarially chosen perturbations of the input.  We hypothesize that this counterintuitive behavior is a result of the high-dimensional geometry of the data manifold, and explore this hypothesis on a simple high-dimensional dataset. For this dataset we show a fundamental bound relating the classification error rate to the average distance to the nearest misclassification, which is \emph{independent of the model}.  We train different neural network architectures on this dataset and show their error sets approach this theoretical bound. As a result of the theory, the vulnerability of machine learning models to small adversarial perturbations is a logical consequence of the amount of test error observed. We hope that our theoretical analysis of this foundational synthetic case will point a way forward to explore how the geometry of complex real-world data sets leads to adversarial examples.
\end{abstract}

\section{Introduction}  Many standard image models exhibit a surprising phenomenon: most randomly chosen images from the data distribution are correctly classified and yet are close to a visually similar nearby image which is incorrectly classified \citep{goodfellow2014explaining, Szegedy14}. This is often referred to as the phenomenon of {\em adversarial examples}. These adversarially found errors can be constructed to be surprisingly robust, invariant to viewpoint, orientation and scale \citep{Athalye17}. Despite some theoretical work and many proposed defense strategies \citep{ pmlr-v70-cisse17a, madry2017advexamples, papernot2016distillation} the cause of this phenomenon is still poorly understood.

Various ideas have been proposed, but no consensus has been reached on an explanation. One common hypothesis is that neural network classifiers are too linear in various regions of the input space, \citep{goodfellow2014explaining, luo2015foveation}. Another hypothesis is that adversarial examples are off the data manifold \citep{Goodfellow-et-al-2016, song2017pixeldefend, samangouei2018defense, lee2017generative}.
\citep{pmlr-v70-cisse17a} argue that large singular values of internal weight matrices may cause the classifier to be vulnerable to small perturbations of the input. 

These explanations assume the presence of adversarial examples stems from some sort of flaw in model architecture or training procedure. Motivated by potential security concerns, many authors have sought ways to defend against this phenomenon. Some works increase robustness to small perturbations by changing the model architecture \citep{krotov2017dense}, distilling a large network into a small network \citep{papernot2016distillation}, using regularization \citep{pmlr-v70-cisse17a}, or detecting adversarial examples statistically \citep{feinman2017detecting, abbasi2017robustness, grosse2017statistical, metzen2017detecting}. Many of these methods, however, have been shown to fail \citep{carlini2017adversarial, carlini2017towards, obfuscated-gradients}. A strong recent empirical result uses adversarial training to increase robustness \citep{madry2017advexamples}. However, local errors have still been shown to appear for distances just beyond what is adversarially trained for \cite{chen2017madry}. 

In this paper we investigate a new potential explanation for adversarial examples, which may provide insight into the difficulties faced by these defenses. We hypothesize that this behavior is due to the high-dimensional geometry of data manifolds combined with the presence of low but non-zero error rates. In order to begin to investigate this hypothesis, we define a simple but foundational task of classifying between two concentric $n$-dimensional spheres. This allows us to study adversarial examples in a simplified setting where the data distribution $p(x)$ is well defined mathematically and where we have an analytic characterization of the decision boundary learned by the model. 

As we study this dataset we are interested in the relationship between two fundamental measures about the error set $E$. First is the measure under the data distribution of the error set $\mu(E) = \mathbb{P}_{x\sim p}[x \in E]$. The second is the average distance to the nearest error $\de = \mathbb{E}_{x \sim p} d(x, E)$ where $d(x,E)$ denotes the $l_2$ distance from $x$ to the nearest point in the error set $E$. The phenomenon of adversarial examples can be stated as the surprising fact that for very accurate models ($\mu(E)$ is small), most points in the data distribution are very close to an error ($\de$ is also small). We consider the average distance for simplicity. The quantity $\de$ is related to ``adversarial robustness'' often studied in adversarial defense papers, which estimates $\mathbb{P}_{x \sim p} [d(x,E) < \epsilon]$\footnote{In our experiments, we observed that the distribution of $d(x,E)$ is tightly concentrated (Appendix A). In such cases $\de$ will approximate the largest $\epsilon$ for which a model can exhibit significant $l_2$ adversarial robustness.}.

 Our experiments and theoretical analysis of this dataset demonstrate that the quantities $\mu(E)$ and $d(E)$ can behave in counter-intuitive ways when the data manifold is high-dimensional. In particular we show the following:

\begin{itemize}
    \item Several models trained on this dataset are vulnerable to adversarial examples, that is most randomly chosen points from the data distribution are correctly classified and yet are "close" to an incorrectly classified input. This behavior occurs even when the test error rate is less than 1 in 10 million and even when we restrict the adversarial search space to the \textit{data manifold}.
    \item For this dataset, we prove a fundamental upper bound on $d(E)$ in terms of $\mu(E)$. In particular, we prove that \emph{any} model which misclassifies a small constant fraction of the data manifold will be vulnerable to adversarial perturbations of size $O(1/\sqrt{n})$.
    \item Neural networks trained on this dataset approach this theoretical optimal bound. This implies that in order to linearly increase $d(E)$, the classification error rate $\mu(E)$ must decrease significantly.
 
\end{itemize}

The upper bound we prove has implications for potential defenses against adversarial examples. In particular, for this dataset, the only way to significantly increase adversarial robustness is to reduce test error. We conclude with a discussion about whether or not a similar conclusions hold for other datasets.

\subsection{Related Work}
The relationship between adversarial examples and high-dimensional geometry has also been explored in some prior work. For example \cite{fawzi2018noise} show that, under a locally flat assumption of the decision boundary, distance to the decision boundary is related to model generalization in the presence of Gaussian noise added to the input. \cite{schmidt2018adversarially} prove that increasing adversarial robustness requires more data. Our results are consistent with the theoretical results in \cite{schmidt2018adversarially}, in particular we observe that given enough data and a proper model class, it is possible to remove adversarial examples on this dataset. 

Another impossibility theorem for a synthetic dataset was shown in \cite{fawzi2018adversarial}. This impossibility theorem implies that \emph{any} classifier on the considered dataset (even perfect classifiers) is vulnerable to adversarial perturbations of a certain form. The goal of our work is to not exhibit a dataset which is difficult to solve, but to instead understand the relationship between error rates and the presence of adversarial examples. To our knowledge our work is the first to bound $l_2$-robustness in terms of out-of-sample error rates.


\section{The Concentric Spheres Dataset}

The data distribution consists of two concentric spheres in $n$ dimensions: we generate a random $x \in \mathbb{R}^n$ where $||x||_2$ is either $1.0$ or $R$, with equal probability assigned to each norm (for this work we choose $R = 1.3$). We associate with each $x$ a label $y$ such that $y = 0$ if $||x||_2 =1$ and $y=1$ if $||x||_2 = R$.

Studying a synthetic high-dimensional dataset has many advantages:
\begin{itemize}
    \item The probability density of the data $p(x)$ is well defined and is uniform over all $x$ in the support. 
    \item We can also sample uniformly from $p(x)$ by sampling $z \sim N(\vec{0}, I)$ and then setting $x = z/||z||_2$ or $x = Rz/||z||_2$. 
    \item We can design machine learning models which can provably learn a decision boundary which perfectly separate the two spheres.
    \item We can control the difficulty of the problem by varying the input dimension $n$.
\end{itemize} 

Our choice of $R = 1.3$ was arbitrary and we did not explore in detail the relationship between adversarial examples and the distance between to two spheres. Additionally, our choice to restrict the data distribution to be the shells of the two spheres was made to simplify the problem further.

We denote $N$ to be the size of the training used. For some experiments we train in the online setting, where each mini-batch is an iid sample from $p(x)$ (that is $N=\infty$). All models were trained to minimize the sigmoid cross-entropy loss $L(\theta, x, y)$ where $\theta$ denotes the model parameters.

\subsection{Finding Adversarial Examples on the Data Manifold} \label{sec:manifold}
Several recent works have hypothesized that adversarial examples are off the data manifold and designed defenses based on this hypothesis \citep{song2017pixeldefend, samangouei2018defense, lee2017generative}. In order to test this hypothesis on this dataset we designed an attack which specifically produces adversarial examples on the data manifold which we call the {\em manifold attack}. 

Let $(x,y)$ denote an example from the data distribution. To find adversarial examples $\hat{x}$ on the data manifold we maximize $L(\theta, \hat{x}, y)$ subject to the constraint $||\hat{x}||_2 = ||x||_2$. This ensures that the produced adversarial example is of the same class as the starting point and lies in the support of the data distribution. We solve this constrained optimization problem by running 1000 steps of projected gradient descent (PGD) with step size .01, only for the projection step we project back on the sphere by normalizing $||\hat{x}||_2$. To find nearby adversarial examples we terminate PGD at the first error found. Note that finding the nearest error is in general NP-hard \cite{katz2017reluplex}, so we rely on this standard attack technique in order to estimate $d(E)$. Sometimes we are interested in \emph{worst-case examples}, where we iterate PGD for all 1000 steps, potentially walking very far along the sphere from the starting point $x$.

\section{Adversarial Examples for a Deep ReLU Network}\label{sec:relu}

\begin{figure*}
\centering



\includegraphics[width=1.0\linewidth]{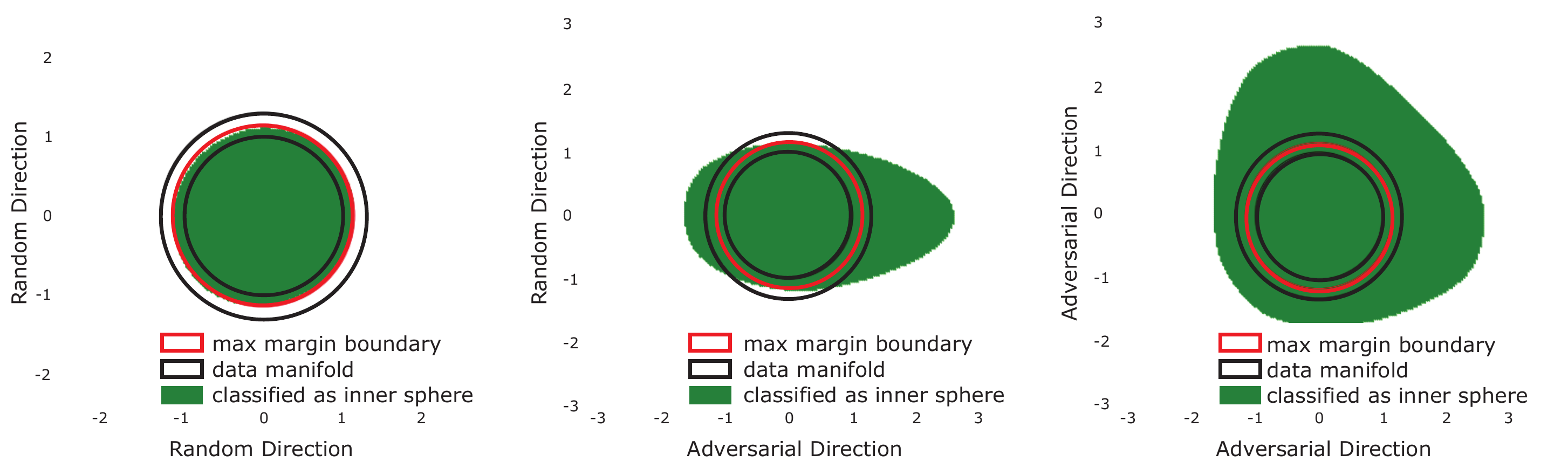}

\caption{Visualizing a 2D slice of the input space where the subspace is spanned by: 2 randomly chosen directions \textbf{(left)}, 1 random and 1 "adversarial direction" \textbf{(center)}, and 2 orthogonal "adversarial directions" \textbf{(right)}. The data manifold is indicated in black and the max margin boundary in red. The green area indicates points which are classified by the ReLU network as being on the inner sphere. In the last plot, the slice of the entire outer sphere is misclassified despite the fact that the error rate of the model is less than 1 in 10 million.}
\label{fig:2dslice}
\end{figure*}

Our first experiment used an input dimensionality of $n = 500$. We then train a 2-hidden-layer ReLU network with 1000 hidden units on this dataset. We applied batch normalization \citep{icml2015_ioffe15} to the two hidden layers, but not to the readout layer. We train with mini-batch SGD, minimizing the sigmoid cross entropy loss. We use the Adam optimizer \citep{Kingma14} for 1 million training steps with mini-batch size 50 and learning rate 0.0001. Thus, over training 50 million unique data points were seen.

We evaluated the final model on 10 million uniform samples from each sphere --- 20 million points in total --- and observed no errors. Thus the error rate of this model is unknown, we only have a statistical upper bound on the error rate. Despite this, we are able to adversarially find errors {\em on the data manifold} with the manifold attack.

In Figure~\ref{fig:2dslice} we investigate the decision boundary by visualizing different 2D subspaces of the 500 dimensional space. Note that the intersection of the two spheres on a 2D subspace will be two concentric circles. When we take a random slice, we see the model has closely approximated the max margin boundary on this slice. By contrast, when we take a 2D slice where one basis vector is a worst-case adversarial example, the model's decision boundary is highly warped along this "adversarial direction". There are points of norm greater than $2$ that the model confidently classifies as being on the inner sphere. We can also take a slice where the x and y axis are an orthogonal basis for the subspace spanned to two different worst-case examples. Although the last plot shows that the entire slice of the outer sphere is misclassified, the volume of this error region is exceedingly small due to the high-dimensional space.

Despite being extremely rare, these misclassifications appear close to randomly sampled points on the sphere. We measured the mean $l_2$ distance to the nearest error on the inner sphere to be $0.18$. This is significantly smaller than the distances between random samples from $p(x)$. In particular, when we sampled 10000 random points on the inner sphere, the nearest pair was distance 1.25 away from each other.

This phenomenon only occurs in high-dimensions. We found that the same model trained on 100 samples with $n=2$ makes no errors on the data manifold. We also explored the above training procedure for $n \in \{10, 20 ,30, 40, 50, 60, 70, 80, 90, 100\}$ and found that when $n < 100$ the attack algorithm fails to find errors (this of course does not imply that no adversarial examples exists). Additional details can be found in Appendix B.

\section{Analytic Forms for a Simpler Network}

\label{sec:quadratic}

It has been hypothesized that the existence of adversarial examples may be a result of some sort of flaw in model architecture. For example, \cite{krotov2017dense} hypothesized that changing the non-linearity of neural networks may improve robustness and \cite{papernot2016distillation} tried increasing robustness by distilling a large neural network into a smaller one\footnote{This method was later shown to not increase robustness \cite{carlini2017adversarial}.}. To investigate the role of architecture we study a simpler model which intuitively is better suited for this problem. The network, dubbed ``the quadratic network,'' is a single hidden layer network where the pointwise non-linearity is a quadratic function, $\sigma(x) = x^2$. There is no bias in the hidden layer, and the output simply sums the hidden activations, multiplies by a scalar and adds a bias. With hidden dimension $h$ the network has $n \cdot h + 2$ learnable parameters. The logit is written as
\begin{equation} \label{eq:quad_defn}
    \hat{y}(x) = w \vec{1}^T (W_1 x)^2 + b
\end{equation}
 where $W_1 \in \mathbb{R}^{h \times n}$, $\vec{1}$ is a column vector of $h$ 1's. Finally, $w$ and $b<0$ are learned scalars. In Appendix C, we show that the output of this network can be rewritten in the form 
\begin{equation} \label{eq:the quadratic network}
    \hat{y}(x) = \sum\limits_{i = 1}^d \alpha_i z_i^2 - 1
\end{equation}
where $\alpha_i$ are scalars which depend on the model's parameters and the vector $\vec{z}$ is a rotation of the input vector $\vec{x}$. The decision boundary of the quadratic network is all inputs where $\sum\limits_{i=1}^n \alpha_i z_i^2 = 1$. It is an $n$-dimensional ellipsoid centered at the origin. This allows us to analytically determine when the model has adversarial examples. In particular, if there is any $\alpha_i > 1$, then there are errors on the inner sphere, and if there are any $\alpha_i < 1/R^2$, then there are errors on the outer sphere. Therefore, the model has perfect accuracy if and only if all $\alpha_i \in [1/R^2, 1]$. 

When we train the quadratic network with $h=1000$ using the same setup as in Section~\ref{sec:relu} we arrive at the perfect solution: all of the $\alpha_i \in [1/R^2, 1]$ and there are no adversarial examples. If we instead limit the size of the training set to $N=10^{6}$ we arrive at a model which empirically has a very low error rate --- randomly sampling 10 million data points from each sphere results in no errors --- but for which there are adversarial examples. In fact, 325 out of 500 of the learned $\alpha_i$ are incorrect in that $\alpha_i \not\in [1/R^2, 1]$ (for a complete histogram see Figure~\ref{fig:quad}). This means that the ellipsoid decision boundary is badly warped in many of the dimensions.

We can use the Central Limit Theorem (CLT) to estimate the error rate of the quadratic network from the $\alpha_i$ (Section~\ref{sec:clt}). The estimated error rate of this particular model is $\approx 10^{-20}$.

Next we augmented the above setup with a ``perfect'' initialization; we initialize the quadratic network at a point for which all of the $\alpha_i$ are ``correct'' but there are non-zero gradients due to the sigmoid cross-entropy loss. The network is initialized at a point where the sigmoid probability of $y=1$ for the inner sphere and outer spheres is $.0016$ and $0.9994$ respectively. As shown in Figure~\ref{fig:quad} continued training from this initialization results in a rapid divergence of the worst-case and average-case losses. Although the average loss on the test set decreases with further training, the worst-case rapidly increases and adversarial examples can once again be found after 1000 training steps. This behavior results from the fact that the training objective (average sigmoid cross-entropy loss) does not directly track the accuracy of the models. 

\subsection{Analytic Estimates of the Error Rate} \label{sec:clt}

\begin{figure*}


\makebox[\textwidth][c]{\includegraphics[width=1.03\linewidth]{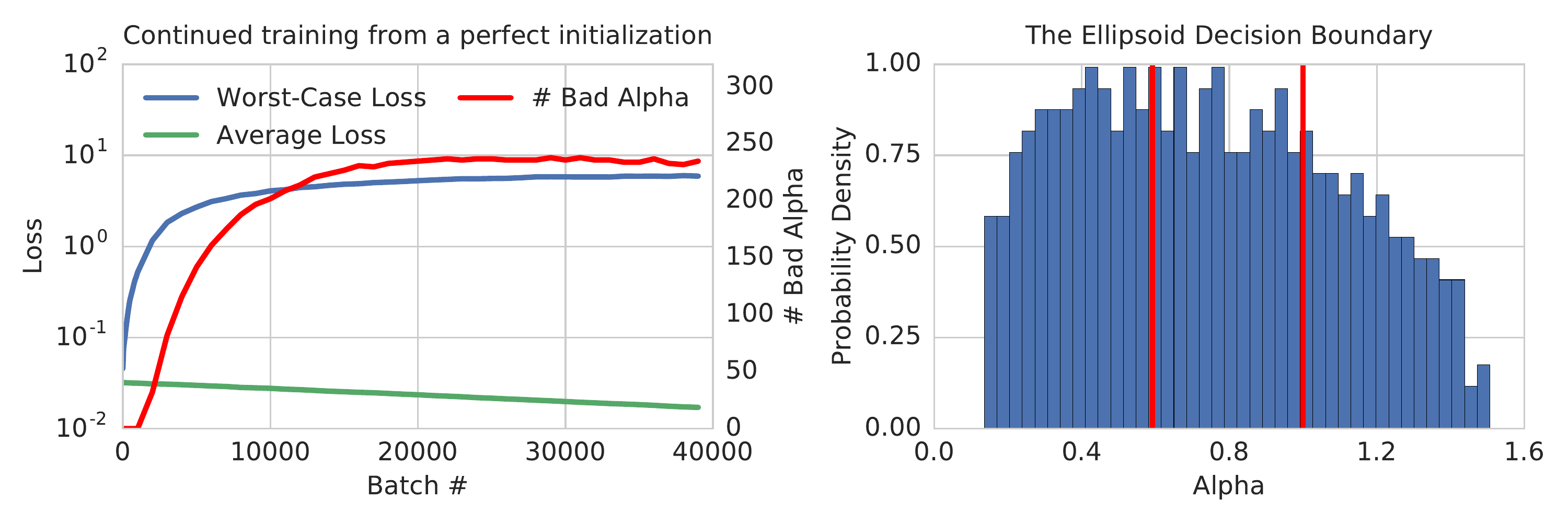}}%

\caption{
\textbf{Left:} The final distribution of $\alpha_i$ when the quadratic network is trained on $10^6$ examples. In order for the model to have $\mu(E) = 0$, all $\alpha_i$ must lie between the two red lines. Despite an estimated error rate of $10^{-11}$, 394 out of 500 of the $\alpha_i$ are incorrect. This means the ellipsoid decision boundary is badly warped along most axes. \textbf{Right:} Training curves of the quadratic network initialized at a perfect initialization with no classification errors. As training progresses average loss gets minimized at the cost of a dramatically worse worst-case loss. The number of incorrect $\alpha_i$ increases at a similar rate.}
\label{fig:quad}
\vspace{-.5cm}
\end{figure*}



In Appendix D, we show how to use the CLT to analytically estimate the accuracy for the quadratic network in terms of the $\alpha_i$. In particular, let $\mu = \sum\limits_{i=1}^d (\alpha_i - 1)$ and $\sigma = 2 \sum\limits_{i=1}^d (\alpha_i - 1)^2$. Then the error rate on the inner sphere can be estimated as
    \begin{equation} \label{eq:approx}
        \mathbb{P}_{\vec{x} \sim S_0} \left[ \sum\limits_{i=1}^d \alpha_i x_i^2 > 1\right]  \approx 1 - \Phi\left(\frac{\mu}{\sigma}\right).
    \end{equation}

  We consider the Gaussian approximation\footnote{Standard Chernoff bounds are also too loose for the models we consider.} as it yields a simple, easy to understand formula, and obtaining exact values is not necessary for the purposes of this section.

Equation~\ref{eq:approx} implies that there are many settings of $\alpha_i$ which obtain very low but non-zero error rates. As long as $\frac{1}{n} \sum\limits_{i=1}^n \alpha_i \approx (1 + R^{-2})/2$ and the variance is not too high, the model will be extremely accurate. The histogram in Figure~\ref{fig:quad} illustrates this; i.e. the learned model has an estimated error rate of $10^{-11}$ but 80\% of the $\alpha_i$ are incorrect. \textbf{For a typical sample, the statistical model sums ``incorrect'' numbers together and obtains the correct answer.} Flexibility in choosing $\alpha_i$ while maintaining good accuracy increases dramatically with the input dimension. This principal may carry over to image datasets, there may be many ways for a convolutional neural network to learn correlations between input features and the output that result in impressive generalization but do not involve a human level understanding of task at hand \cite{jo2017measuring}. We discuss this further in Appendix E.

\section{A Small Amount of Classification Error Implies Local Adversarial Examples}

The theoretical analysis of the quadratic network in Section~\ref{sec:clt} explains why $\mu(E)$ might be very small but non-zero. In this section we prove a fundamental bound on the quantity $d(E)$ in terms of $\mu(E)$.  

Let $S_0$ be the sphere of radius 1 in $n$ dimensions and fix $E \subseteq S_0$ (we interpret $E$ to be the set of points on the inner sphere which are misclassified by some model).  We prove the following theorem in Appendix F:

 \begin{theorem} \label{thm:close}
    Consider any model trained on the sphere dataset. Let $p \in [0.5,1.0)$ denote the accuracy of the model on the inner sphere, and let $E$ denote the points on the inner sphere the model misclassifies (so in measure $\mu(E) = 1-p$). Then $d(E) \leq O(\Phi^{-1}(p)/\sqrt{n})$ where $\Phi^{-1}(x)$ is the inverse normal cdf function.
 \end{theorem}
 
\begin{figure*}
\centering
  \includegraphics[width=1.0\linewidth]{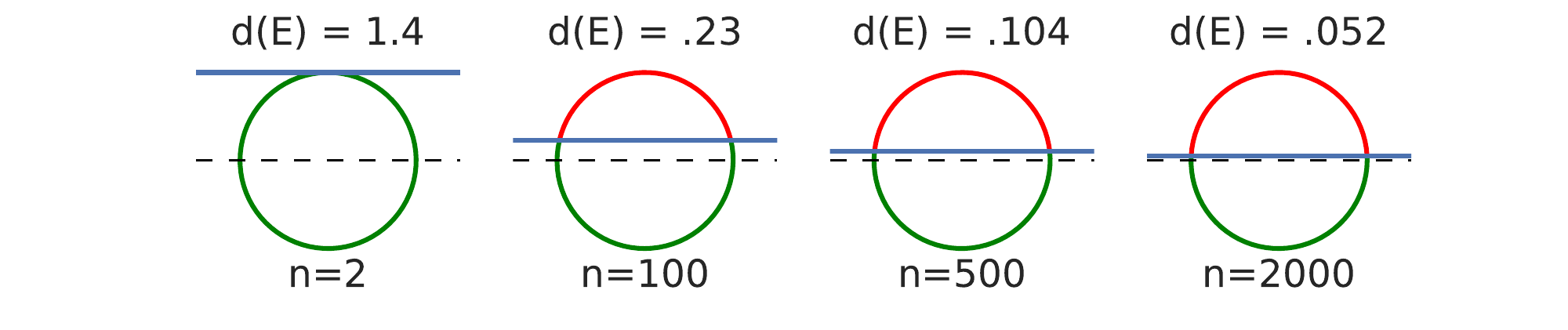}
\caption{We visualize Theorem 5.1 as the dimension of the inner sphere increases. In each plot we visualize an error set of measure $\mu(E) = .01$ (indicated in red) which achieves the theoretically optimal average adversarial distance $d(E)$. As the dimension increases the error set (which is a spherical cap) extends all the way to the equator. This means that $d(E)$ shrinks despite the fact that $\mu(E)$ is held fixed. Note that the mass of an $n$-dimensional sphere is concentrated near the equator. This allows the measure of the error set to be small despite being so close to the equator.}
\label{fig:visual}

\vspace{-.5cm}
\end{figure*}

 This theorem directly links the probability of an error on the test set to the average distance to the nearest error {\em independently of the model}. Any model which misclassifies a small constant fraction of the sphere must have errors close to most randomly sampled data points, no matter how the model errors are distributed on the sphere. At a high level it follows as direct corollary of an isoperimetric inequality of~\cite{figiel1977dimension}. The error set $E$ of fixed measure $\mu(E)$ which maximizes the average distance to the nearest error $d(E)$ is a "cap", which is a set of the form $E = \{x \in S_0 : \bm{w} \cdot \bm{x} > b\}$ (the sphere intersected with a halfspace). This is visualized for different dimensions in Figure~\ref{fig:visual}. Even if one were to optimally distribute the error set $E$ of measure $\mu(E)$ in a local region near a pole of the sphere, this region must extend all the way to the equator and $d(E)$ would be small. 
 
\begin{figure}
\centering
  \includegraphics[width=0.7\linewidth]{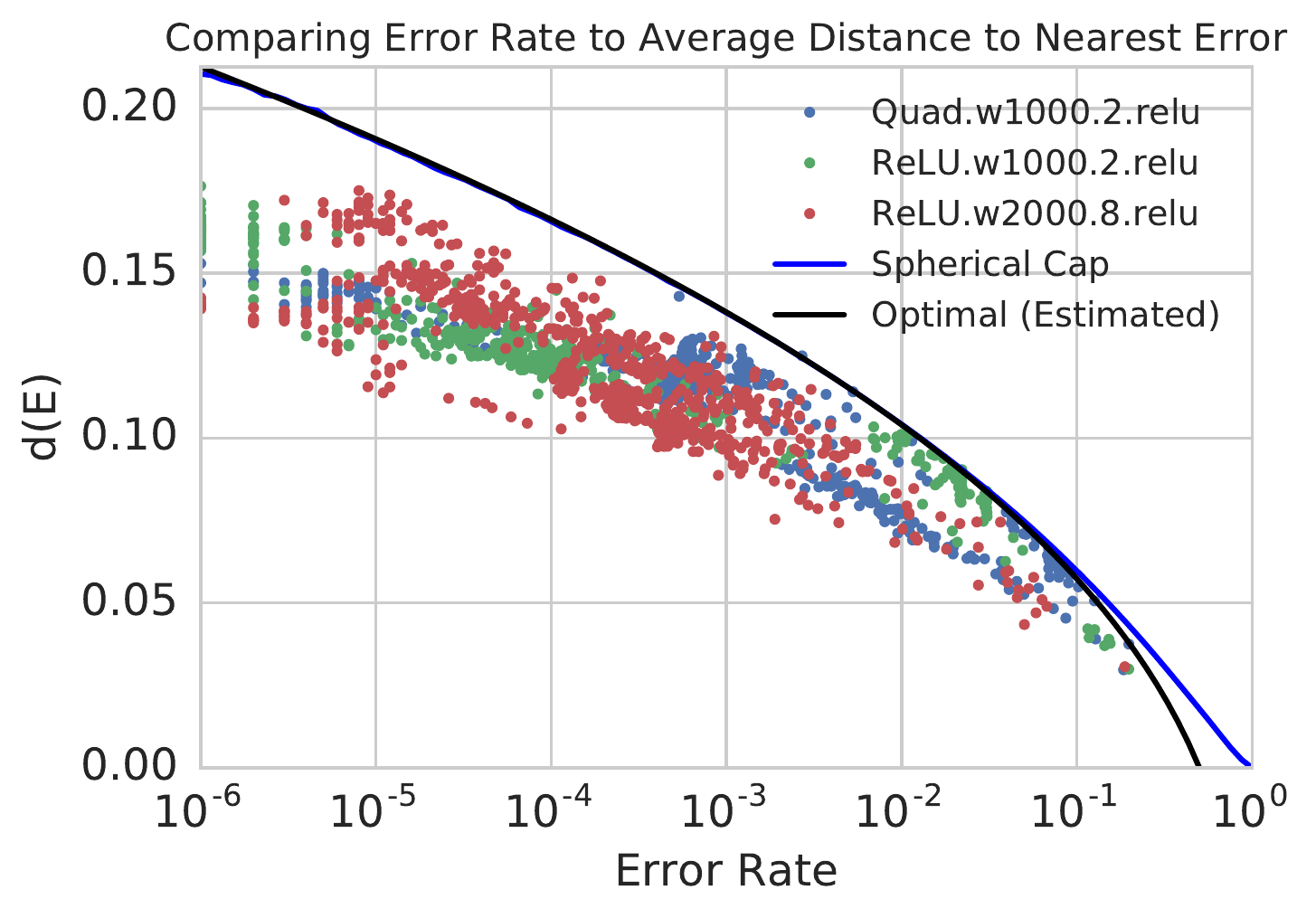}
\caption{We compare the average distance to nearest error with error rate for 3 networks trained on the sphere dataset. All errors are reported for the inner sphere. The 3 networks are trained with 5 different training set sizes, and their performance are measured at various points during training (the networks eventually become too accurate to appear on this graph, as the error rate will be too small to estimate from a random sample. Surprisingly, our empirical measurements of $d(E)$ for these different models are all close to optimal given the measure of the error set $\mu(E)$. We plot the optimal curve in two ways, first with empirical measurements of actual spherical caps (blue), and second with the approximate bound given in Theorem~\ref{thm:close} (black). Note that due to some noise in estimating the error rates and average distances, some networks may appear slightly better than optimal. }
\label{fig:amazing}

\vspace{-.5cm}
\end{figure}

 Theorem~\ref{thm:close} gives an optimal bound on $d(E)$ in terms of $\mu(E)$. In Figure~\ref{fig:amazing}, we can compare how the error sets of actual trained neural networks compare with this optimal bound. We train three different architectures on the sphere dataset when $d=500$, the first is a "small" ReLU network with 1000 hidden units and 2 hidden layers (ReLU-h1000-d2), the second is the quadratic network with 1000 hidden units (Quad-h1000), and the third is a "large" ReLU network with 2000 hidden units and depth 8 (ReLU-h2000-d8). We train these networks with varying number of samples from the data distribution, in particular $N\in \{1000,5000,10000,100000,\infty\}$. For each architecture and training set size $N$, a model is trained 4 unique times with independent initializations (this means that $60$ independent models were trained in total). We then sample the performance of the networks several times during training, computing both the error rate of the model and the average distance to nearest error. The error rate is estimated from $10^6$ random samples from the data distribution and the average distance is estimated by 100 random points and running PGD for 1000 steps with step size of .001 (searching only on the data manifold for errors). Each point on the plot is a network at a certain snapshot during training. When $N \geq 10000$ the networks later in training become so accurate that the error rate cannot be estimated statistically from $10^6$ samples, these points do not appear in the graph. We see that the measured $\de$ is always close to what is theoretically optimal given the observed $\mu(E)$. The optimal $\de$ curve is shown in black and is calculated as $\Phi^{-1}(1- \mu(E))/\sqrt{n}$. We also plot in blue the same optimal curve as estimated by sampling $10^6$ points on the sphere, and calculating their average distance to an actual spherical cap (this is to verify that the estimate provided by Theorem~\ref{thm:close} is accurate). Note that there is some noise in estimating the error rates and average distances (for example PGD is not guaranteed to find the closest error). As a result some networks when sampled appear slightly better than optimal. 
 
 This plot suggests that the decision boundaries of these networks are all well behaved given the amount of test error observed. Interestingly, despite vastly different architectures the relationship between $\mu(E)$ and $d(E)$ is similar for all networks. This is evidence that the geometry of their decision boundaries is similar. It is surprising that an 8-layer ReLU network with 29 million parameters would exhibit a similar relationship between $\mu(E)$ and $d(E)$ as the much simpler quadratic network. 
 
\section{Implications for Real World Datasets}

For the idealized sphere data set, Theorem 5.1 has two key implications. First, in this setting, the only way to reduce the frequency of adversarial examples is to reduce generalization error. Second, the same isoperimetric inequality argument implies there is nothing ``special'' about adversarial examples as a class. That is, for any model with reasonable accuracy, most errors are ``adversarial'' relative to some example data point, in the sense that for a typical incorrectly classified point there is a small perturbation that will cause it to be correctly classified. This means that there is no hope of finding some identifying characteristic of an adversarial example that hints that it is the result of tampering or a special construction.

The natural question is whether these two conclusions apply to real-world data sets. While the sphere data set is highly idealized, it is closely related to a standard Gaussian distribution of examples from a given class. Indeed, it seems plausible that a similar theorem might hold for an anisotropic Gaussian distribution with large variances on sufficiently many dimensions.

Early experiments suggest, however, that the relationship between generalization error and frequency of adversarial examples may be more complicated for real world datasets. In the case of the MNIST dataset, there is a gap between the $d(E)$ observed by current state-of-the-art vision models, and any impossibility theorem one could hope to prove that does not make any assumptions about the model. In particular, in Appendix G we construct a hypothetical set $E$ for MNIST which satisfies $\mu(E) = .01$ and $d(E) = 6.59$. This set is constructed by considering a halfspace aligned with a top PCA direction. It does not correspond to the error set achieved by any machine learning model. By comparison a vanilla convolutional neural network we trained satisfies $\mu(E) = .007$ and $d(E) = 1.3$. The existence of this gap is in agreement with recent empirical results using adversarial training which has increased adversarial robustness without reducing test error \cite{madry2017advexamples}. This also demonstrates that there is still room for increased adversarial robustness before out-of-sample error rates in the natural image distribution must also be reduced.

On the other hand, there is some potential supporting evidence from experiments using state-of-the-art computer vision models, which do not generalize perfectly in the presence of large, visually perceptible Gaussian noise added to an image \cite{goodfellow2014explaining}. There may be a relationship between these sampled test errors and imperceptible, adversarially-found errors. Indeed, recent work \cite{fawzi2018noise} has provided evidence that there is a relationship between the distance to the decision boundary and model generalization in the presence of large, visible noise added to the input.

Overall, we believe it would be interesting for future work to investigate these questions in more detail for other data sets. If similar results hold for real world data sets, it would not imply that improving adversarial robustness is impossible, only that doing so would require improved generalization.

\section{Conclusion and Takeaways}
    In this work we hypothesized that the counter-intuitive behavior of adversarial examples could be a naturally occurring result of the geometry of the high-dimensional manifold. To that end we studied a simple dataset of classifying between two high-dimensional concentric spheres. After training different neural network architectures on this dataset we observed a similar phenomenon to that of image models --- most random points in the data distribution are both correctly classified and are close to a misclassified point. We then explained this phenomenon for this particular dataset by proving a theoretical upper bound on the average distance to nearest error, $d(E)$ in terms of test error $\mu(E)$ which is \emph{independent of the model}. We also observed that several different neural network architectures closely match this theoretical bound.
    
    Theorem~\ref{thm:close} is significant because it reduces the question of why models are vulnerable to adversarial examples to the question of why is there a small amount of classification error. This has implications for the adversarial defense literature, as it implies the only way to defend against small perturbations on this dataset is to significantly reduce $\mu(E)$.  
    
    Our results on this dataset provide a counterexample to many previously proposed hypothesis for the existence of adversarial examples. This could explain why many prior adversarial defense proposals based on other hypothesis have later been shown to fail \cite{carlini2017adversarial,obfuscated-gradients,carlini2017towards}. We believe our results on this simple dataset provide a useful starting point for future investigations into the nature of adversarial examples, the ``adversarial subspace'', or potential issues with model decision boundaries. Regarding increasing adversarial robustness, there is reason to be optimistic. In the case of MNIST, we have shown that there is room for further improvement of adversarial robustness given current out-of-sample error rates. Furthermore, even if similar impossibility results do hold for other datasets, they would not imply that defending against adversarial examples is impossible, only that success in doing so would require improved model generalization. 
    


 \section*{Acknowledgments}
Special thanks to Surya Ganguli, Jascha Sohl-dickstein, Jeffrey Pennington, and Sam Smith for interesting discussions on this problem.

\small
\bibliographystyle{unsrt}
\bibliography{sphere}

\appendix
\part*{Appendix}
\setcounter{figure}{0}
\renewcommand\thefigure{\thesection.\arabic{figure}}

\section{Inspecting the Distribution Over Distances}

\begin{figure*}
\begin{subfigure}{.5\textwidth}
  \centering
  \includegraphics[width=1.0\linewidth]{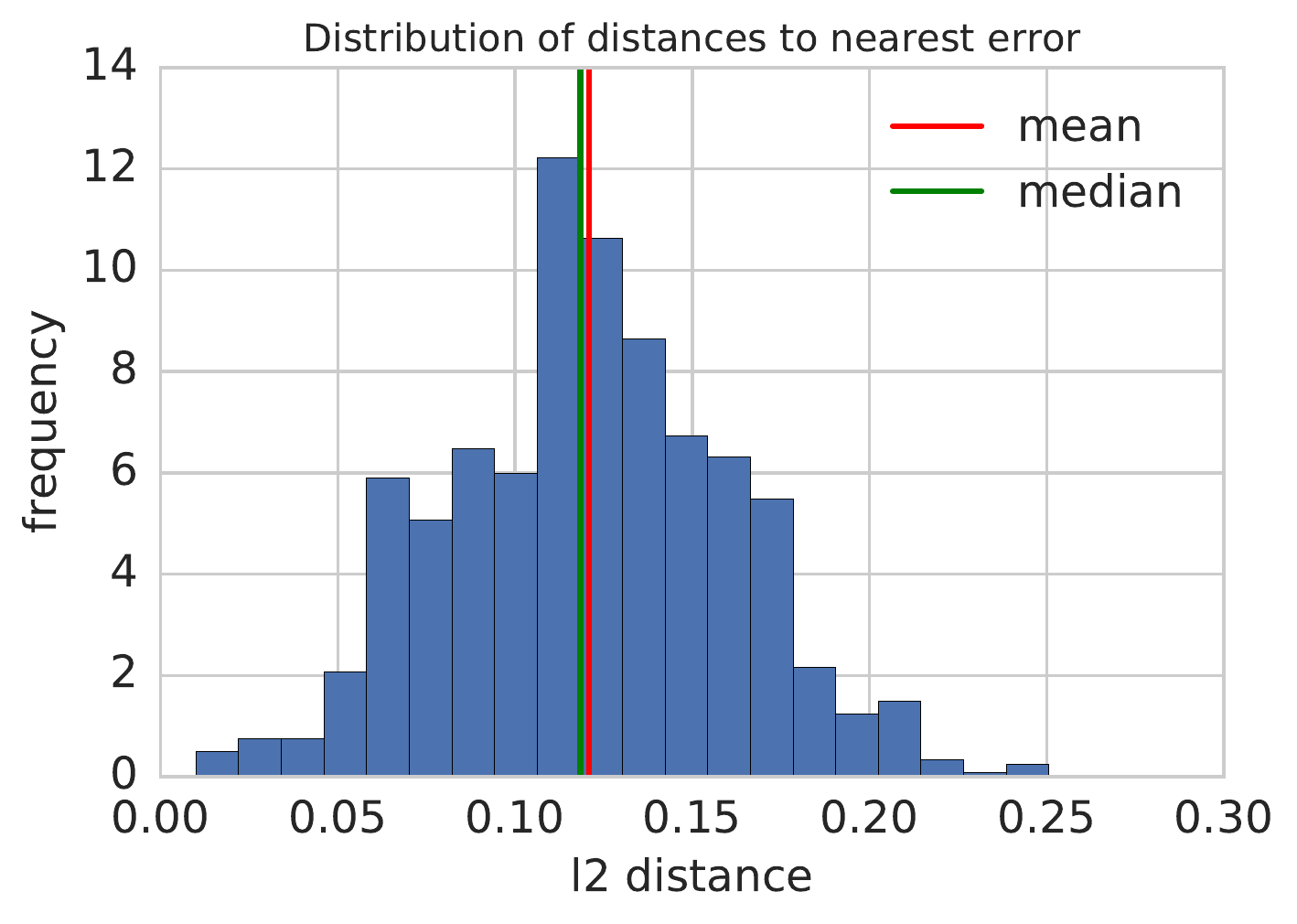}
\end{subfigure}
\begin{subfigure}{.5\textwidth}
  \centering
\includegraphics[width=1.0\linewidth]{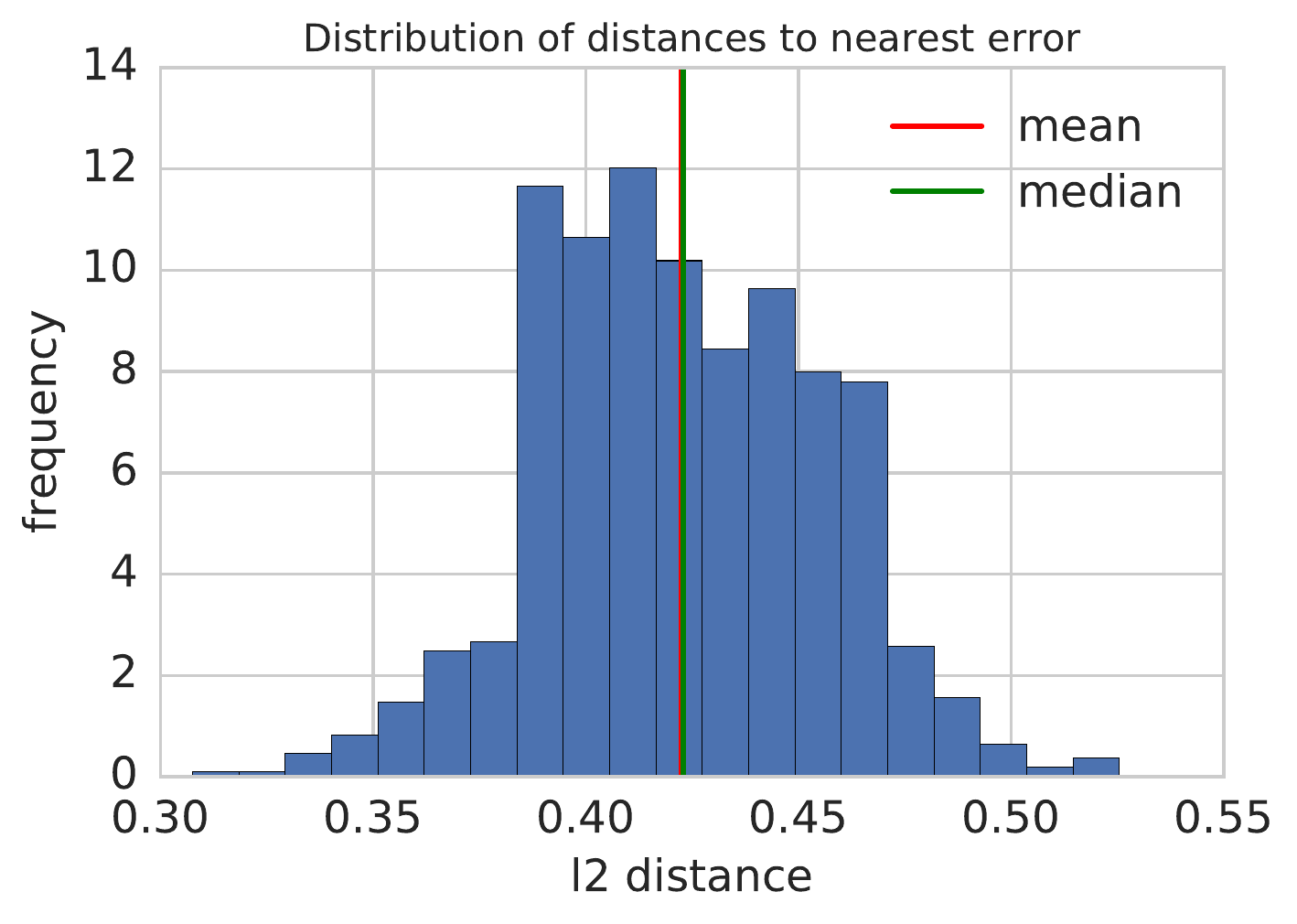}
\end{subfigure}
\caption{
Empirical estimates of the distribution of $d(x, E)$ for different networks. Left: An 8 hidden layer ReLU network with $\mu(E) = .999$. Right: a quadratic network with estimated $\mu(E) = 10^{-20}$.}
\label{fig:dist_distribution}
\vspace{-.5cm}
\end{figure*}

The primary metric considered throughout this work is the average adversarial distance $d(E)$. This metric is considered for simplicity, however we find that it is closely related to the concept of $\epsilon$ $l_2$-robustness typically considered in the adversarial defense literature. Because the distances are tightly concentrated (Figure~\ref{fig:dist_distribution}), we can interpret the mean distance to approximate to the largest $\epsilon$ for which a network is 50\% robust in the $l_2$ metric.

\section{Additional Results on Training}
\begin{figure*}
  \centering
\includegraphics[width=0.7\linewidth]{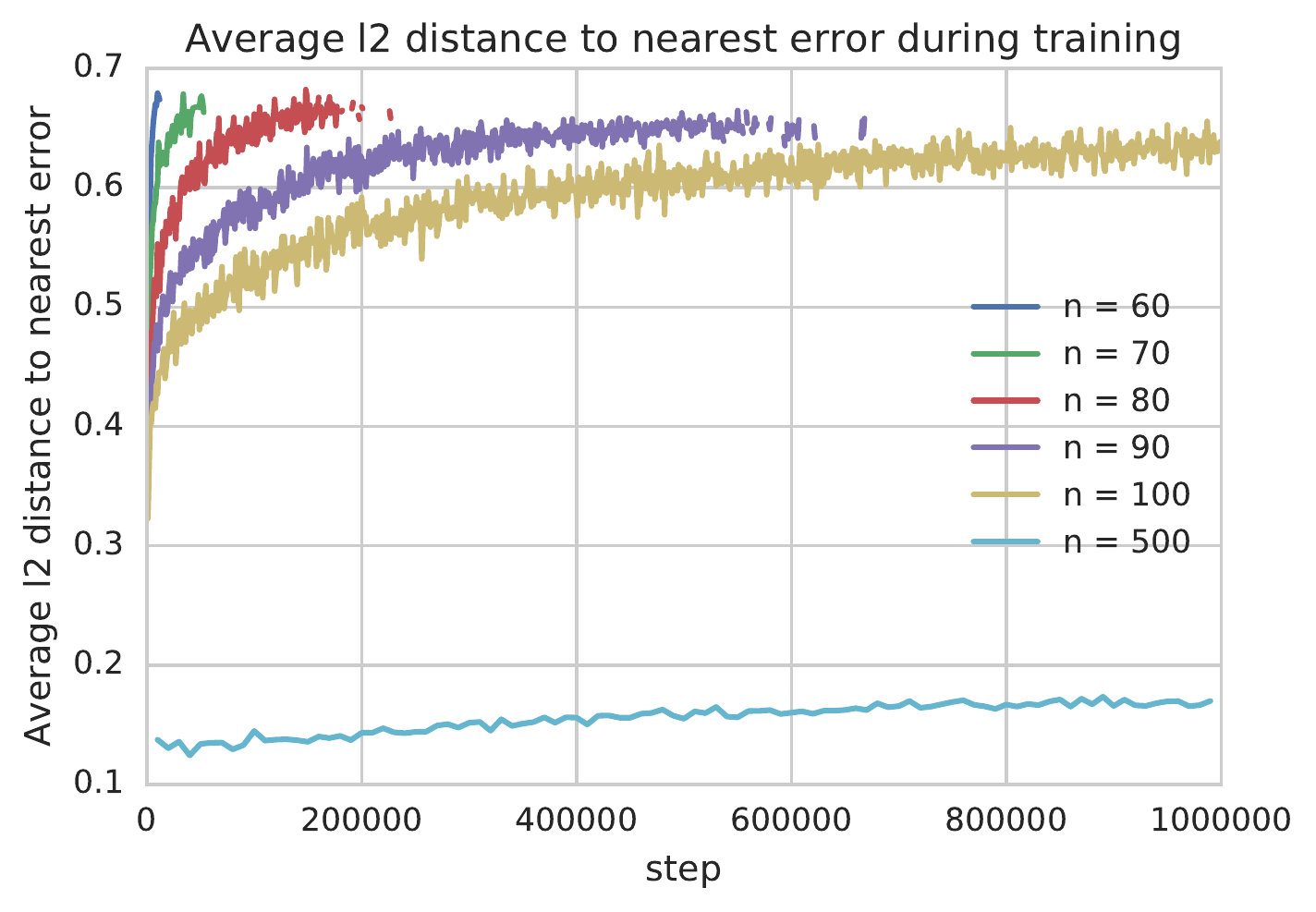}
\caption{
Visualizing $d(E)$ during the course of training. When $n < 100$ the models improve to a point where the attack algorithm is no longer able to find errors (at which case no points are shown).}
\label{fig:input_dim}
\end{figure*}

  We ran the 2-hidden-layer ReLU network for various input dimensions and measured $d(E)$ during training, results are shown in Figure~\ref{fig:input_dim}. $d(E)$ here is estimated from $100$ random starting points on the sphere and running 1000 steps of PGD with step size .01. If no errors are found from these 100 starting points then attack is considered to have failed. When $n < 100$ the models reach a point where the attack algorithm no longer finds errors, no data is plotted for these points. However, when $n=100$ the observed $d(E)$ plateaus around $.7$. For $n=500$, $d(E)$ quickly plateaus at $.18$. Note that a failed attack does not imply that no errors exist. All models quickly become so accurate that test error cannot be estimated statistically.
  
\begin{figure*}
\begin{subfigure}{.5\textwidth}
  \centering
  \includegraphics[width=1.0\linewidth]{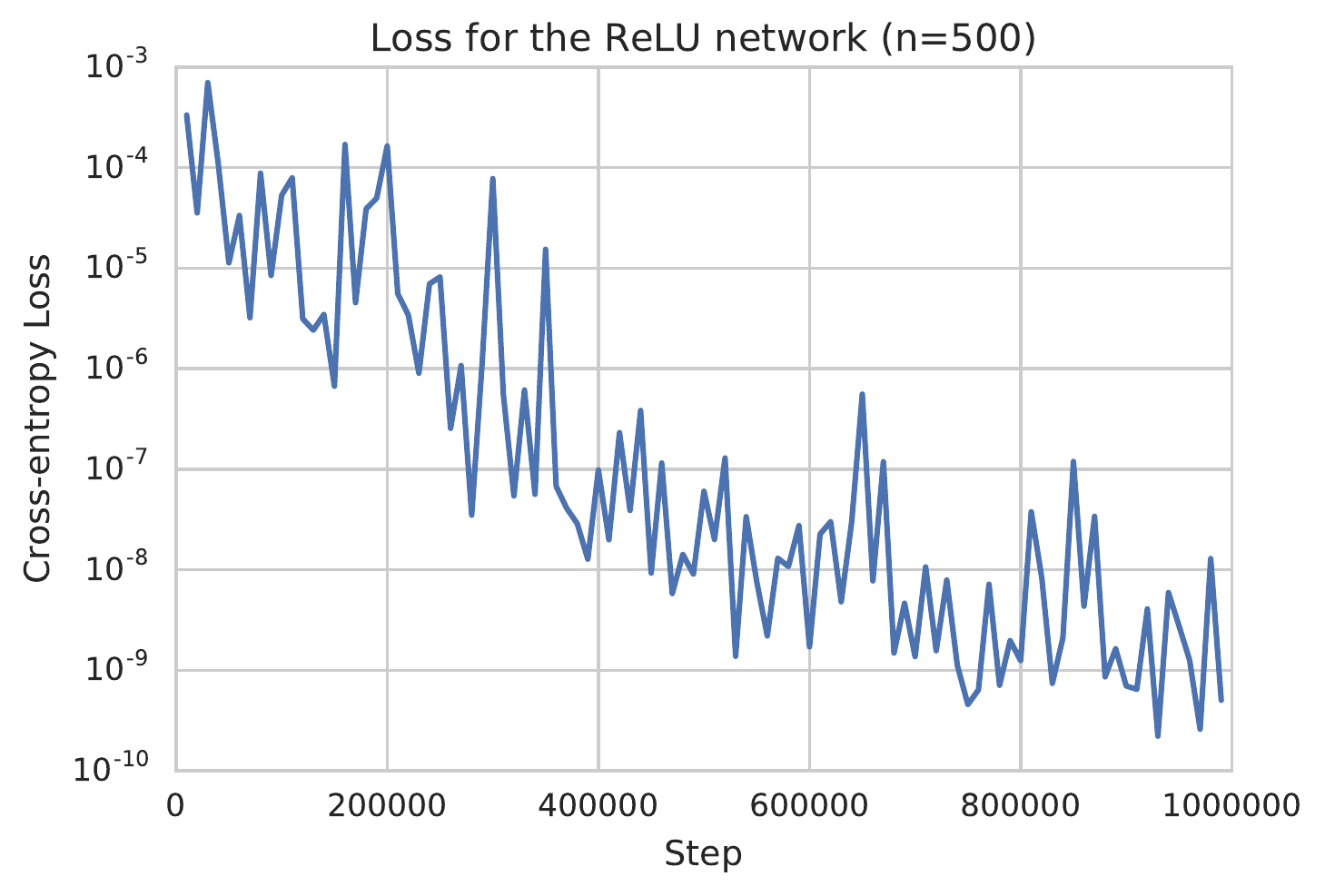}
\end{subfigure}
\begin{subfigure}{.5\textwidth}
  \centering
\includegraphics[width=1.0\linewidth]{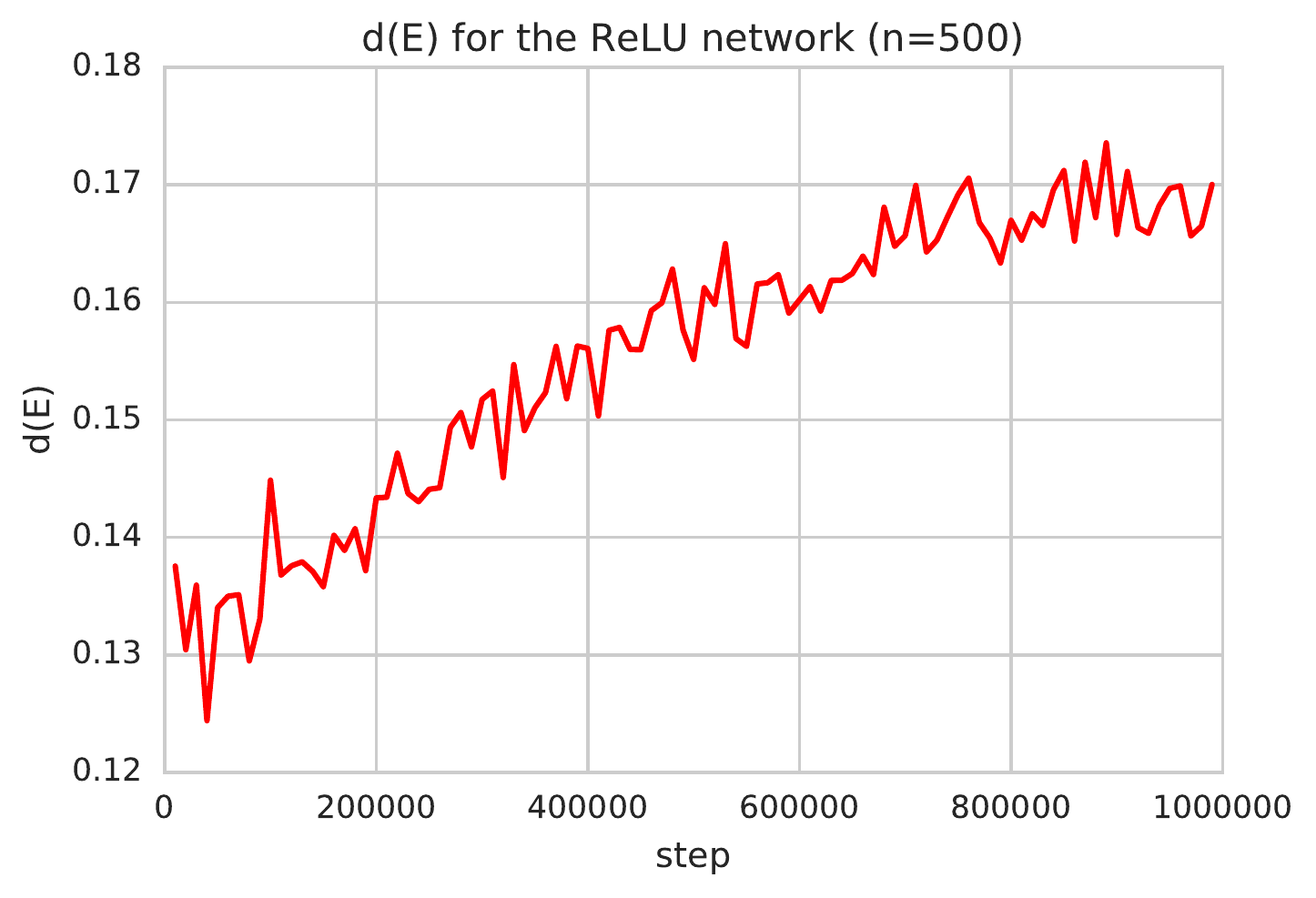}
\end{subfigure}
\caption{
\textbf{Left:} Cross entropy loss for the ReLU network trained on the sphere with $n=500$. Note that the loss is estimated from out of sample data and thus the fact that it is so low ($< 10^{-9}$) is not the result of over-fitting.  \textbf{Right:} $d(E)$ during the course of training. Despite the fact that the loss is so low, there are still errors in the data distribution, and the average distance plateaus around $.18$. }
\label{fig:n500}
\vspace{-.5cm}
\end{figure*}

  In Figure~\ref{fig:n500} we plot the loss of the 2-layer ReLU network trained with $n=500$. The cross-entropy loss (as measured with out-of-sample data) becomes less than $10^{-9}$. The quantity $d(E)$ plateaus around $.17$. 

\section{The Decision Boundary of the Quadratic Network}
Here we show the decision boundary of the quadratic network is a $n$-dimensional ellipsoid.
We'll use $\bm{x}$ to denote the column vector representation of the input. The logit of the quadratic network is of the form 
\begin{equation}
   l(x) =  (W_1 \bm{x})^T (W_1 \bm{x}) w + b
\end{equation}

$W_1$ is the input to hidden matrix, and $w$ and $b$ are scalars. We can greatly simplify this by taking SVD of $W_1 = U \Sigma V^T$. So we have the following:

\begin{align*}
    l(x) & = (W_1 \bm{x})^T (W_1 \bm{x}) w + b \\
    & = (U \Sigma V^T \bm{x})^T (U \Sigma V^T \bm{x}) w + b \\
    & = (\bm{x}^T V \Sigma U^T U \Sigma V^T \bm{x}) w + b
\end{align*}

Let $\bm{z}$ = $V^T \bm{x}$ which is a rotation of the input. Then the above becomes

\begin{equation}
    l(x) =  \bm{z}^T \Sigma^2 \bm{z} w + b
\end{equation}

Letting the singular values of $W_1$ be the sequence $(s_i)$ we have

\begin{equation}
    l(x) = w \sum\limits_{i = 1}^n s_i^2 z_i^2 + b
\end{equation}
The decision boundary is of the form $l(x) = 0$, thus we have

\begin{align*}
    w \sum\limits_{i = 1}^n s_i^2 z_i^2 + b & = 0 \\
    \sum\limits_{i = 1}^n \alpha_i z_i^2 - 1 & = 0
\end{align*}
where $\alpha_i = w * s_i^2 / (-b)$.

Note the distribution of $\bm{z}$ is the same as $\bm{x}$ (they are rotations of each other) so replacing $z_i$ with $x_i$ above yields a rotation of the decision boundary.

\section{Estimating the Accuracy with the Central Limit Theorem}

Here we show how to estimate the accuracy of the quadratic network with the CLT. Let $\{\alpha_i\}_{i=1}^n$ be nonnegative real numbers and $b>0$. Let $S_0$ be the unit sphere in $n$-dimensions. Suppose $z$ is chosen uniformly on $S_0$. Then we wish to compute the probability that 
\begin{equation}\label{eq:decision_boundary}
\sum_{i=1}^n \alpha_i z_i^2 > 1.
\end{equation}
One way to generate $z$ uniformly on $S_0$ is to pick $u_i\sim N(0,1)$ for $1\leq i\leq n$ and let $z_i = u_i/||u||$. It follows that we may rewrite eq.~\eqref{eq:decision_boundary} as,
\begin{align}
&\frac1{||u||^2}\sum_{i=1}^n \alpha_i u_i^2 > 1\nonumber\\
&\sum_{i=1}^n\alpha_i u_i^2 > \sum_{i=1}^n u_i^2\nonumber\\
&\sum_{i=1}^n(\alpha_i - 1)u_i^2 > 0.
\end{align}

Let $X = \sum_{i=1}^n (\alpha_i - 1)u_i^2$. In the case that $n$ is sufficiently large we may use the central limit theorem conclude that $X\sim N(\mu,\sigma^2)$. In this regime $X$ will be determined exclusively by its first two moments.  We can work out each of these separately,
\begin{align}
&\mu = \mathbb E[X] = \sum_{i=1}^n (\alpha_i - 1)\\
&\sigma^2 = \text{Var}[X] = 2\sum_{i=1}^n(\alpha_i - 1)^2.
\end{align}
It follows that,
\begin{equation} \label{eq:clt2}
P(X>0) = P\left(\sigma Z + \mu > 0\right) = P\left(Z > -\frac\mu\sigma\right) = 1-\Phi\left(-\frac\mu\sigma\right)
\end{equation}
which proves the result.

 \section{A Model Can Be Extremely Accurate While Ignoring Most of the Input}
 
 \begin{figure*}
  \centering
\includegraphics[width=0.5\linewidth]{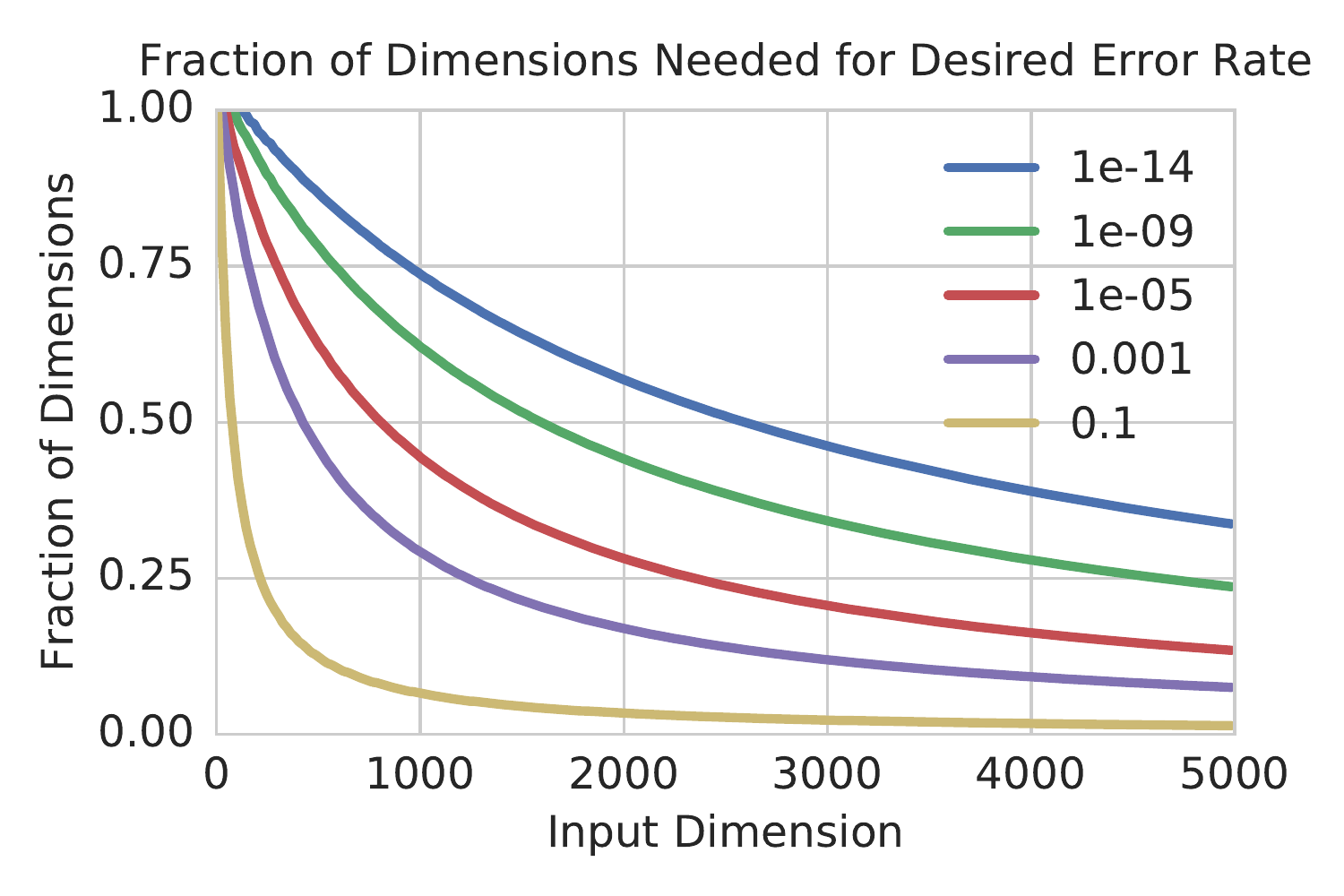}
\caption{
We consider a classification model which only sees a projection of the input, size $n$, onto a $k$ dimensional subspace. We then plot what $k/n$ needs to be in order for the model to obtain a certain error rate. When the input is high dimensional, a model can become extremely accurate despite ignoring most of the input.}
\label{fig:curse}
\end{figure*}
 
Our theoretical analysis of the quadratic network demonstrated that there are many ways for the model to exhibit impressive generalization while fundamentally ``misunderstanding'' the underlying task. That is for a typical example, the quadratic network sums ``incorrect'' numbers together and obtains the correct answer. 

To illustrate this further consider a special case of the quadratic network where the decision boundary is of the form
\[ \sum\limits_{i = 1}^k x_i^2 = b. \]

This simplified model has two parameters, $k$ the number of dimensions the model looks at and $b$ a threshold separating the two classes. How large does $k$ need to be in order for the model to obtain a desired error rate? (Assuming $b$ is chosen optimally based on $k$). We answer this question using the CLT approximation in equation~\ref{eq:clt2}. In Figure~\ref{fig:curse}  we plot the fraction of input dimensions needed to obtain a desired accuracy using this simplified model. For example, if $n = 2000$ then the model can obtain an estimated accuracy of $10^{-14}$ while ignoring 62\% of the input. This behavior happens empirically when we train a quadratic network which is too small to perfectly separate the two spheres. For example, the quadratic network with 1000 hidden nodes trained with $n=2000$ obtains an estimated generalization error of $10^{-21}$ and $10^{-9}$ on the inner and outer spheres respectively despite being only 1/2 the size required to achieve 0 error. This suggests that the size of a network required to achieve very low test error may be significantly smaller than the size of a network required to achieve 0 test error. 

\section{Proof of Theorem~5.1}
\label{appx_alpha0}
In this section we sketch a proof of Theorem~5.1. Let $E$ be the points on the inner sphere $S_0$ which are misclassified by some model. Let $\mu(E)$ denote the measure of $E$ and $d(E) = \mathbb{E}_{x \sim S_0}d(x, E)$ denote the average distance from a random point on $S_0$ to the nearest point in $E$. 

\begin{figure*}
  \centering
\includegraphics[width=0.7\linewidth]{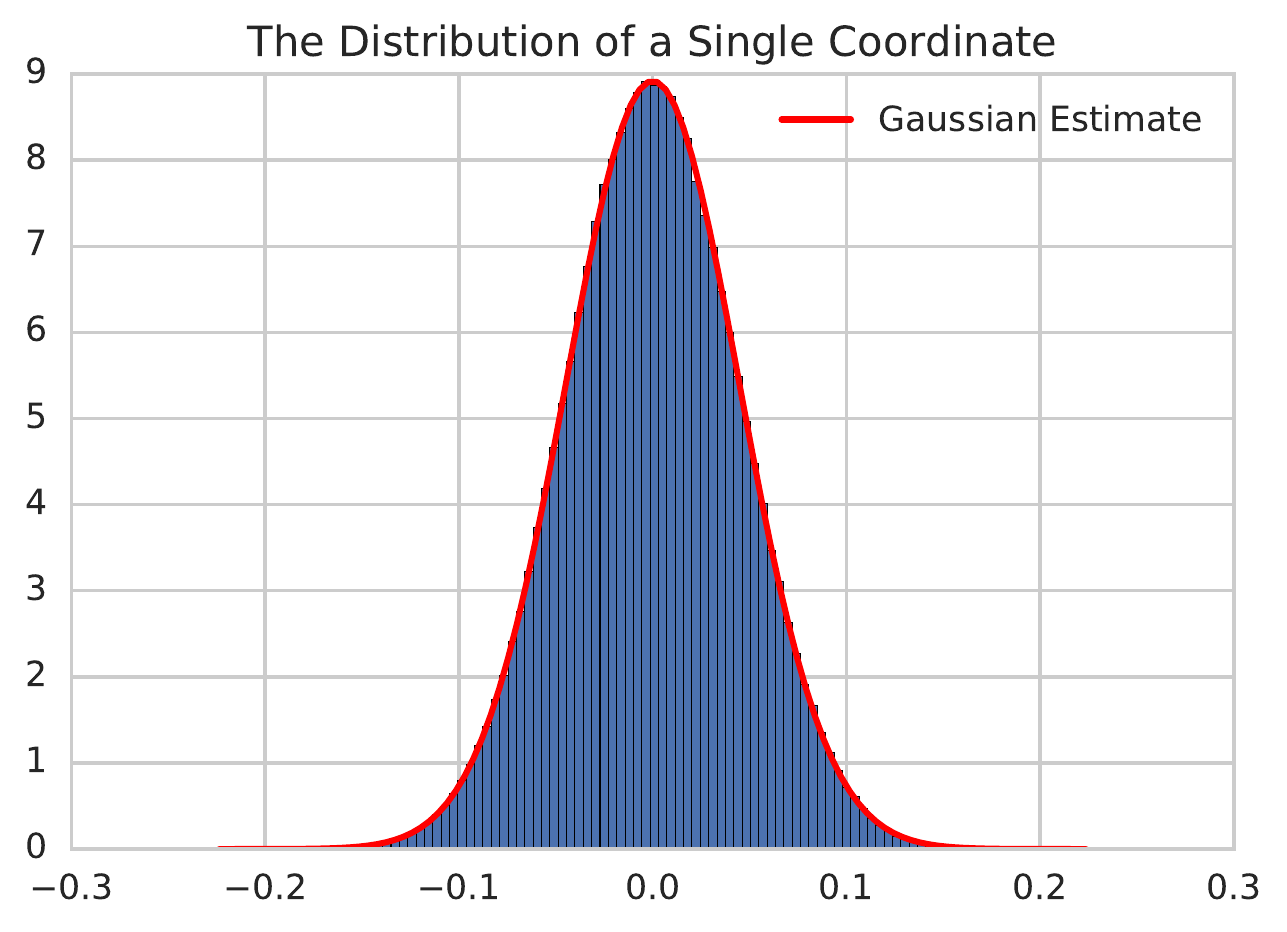}
\caption{ Comparing the distribution of a single coordinate $x_i$ for $x \sim p(x)$ with the Gaussian approximation $N(0, 1/\sqrt{n})$ (here we visualize the case $n=500$).}
\label{fig:single_coordinate}
\end{figure*}

A "cap" of $S_0$ is the intersection of the sphere with a half space of $\mathbb{R}^n$. Intuitively a cap is all points which are located within some fixed distance of a pole. \cite{figiel1977dimension} prove that the set $E$ of given measure $\mu(E)$ which maximizes $d(E)$ is a "cap". Thus, without loss of generality we may assume that $E = \{x \in S_0 : x_1 > \alpha/\sqrt{n}\}$ for some $\alpha > 0$. Obtaining the estimated optimal $d(E)$ thus requires relating the measure of a spherical cap of height $\alpha$ to the average distance from a random sample $x \sim p(x)$ to this cap. Computing this exactly requires solving intractable iterated integrals, so instead we estimate this quantity using a Gaussian approximation. That is, as $n$ becomes large, the distribution of a single coordinate $x_i$ on the sphere approaches $N(0, \frac{1}{n})$. We empirically verify this approximation in Figure~\ref{fig:single_coordinate}. Thus we have

\[\mathbf{P}[x \in E] \approx \mathbf{P}[N(0, \frac{1}{n}) > \alpha/\sqrt{n}] = \mathbf{P}[N(0, 1) > \alpha].\]

Thus we have $\alpha = \Phi^{-1}(\mu(E))$, because this implies $\mathbf{P}[N(0, \frac{1}{n}) > \alpha/\sqrt{n}] = \mu(E)$.

For $x \in S_0$ let $d(x, E)$ denote the distance from $x$ to the set $E$. This distance is equal to $max(\sqrt{2}( \alpha/\sqrt{n} - x_1), 0)$. Thus we have 
\[d(E) = \mathbf{E}_{x \sim S_0} d(x, E) \approx \mathbf{E}\left[max\left(\sqrt{2}(\alpha/\sqrt{n} - N(0, \frac{1}{n})), 0\right)\right] = O(\Phi^{-1}(\mu(E))/\sqrt{n}).\]

We can empirically verify that the quantity $\Phi^{-1}(\mu(E))/\sqrt{n}$ closely approximates the $d(E)$ from actual spherical caps by sampling many points $x$ randomly on the sphere and averaging $d(x,E)$. This is done by sampling $10^{6}$ points, and is used to plot the blue line in Figure 4 of the main text. The black line in the same plot is the function $\Phi^{-1}(\mu(E))/\sqrt{n}$, and closely matches these empirical estimates.

\section{Does a variant of Theorem 5.1 hold on MNIST?}

\begin{figure*}
\centering
\begin{subfigure}{.5\textwidth}
  \centering
  \includegraphics[width=1.0\linewidth]{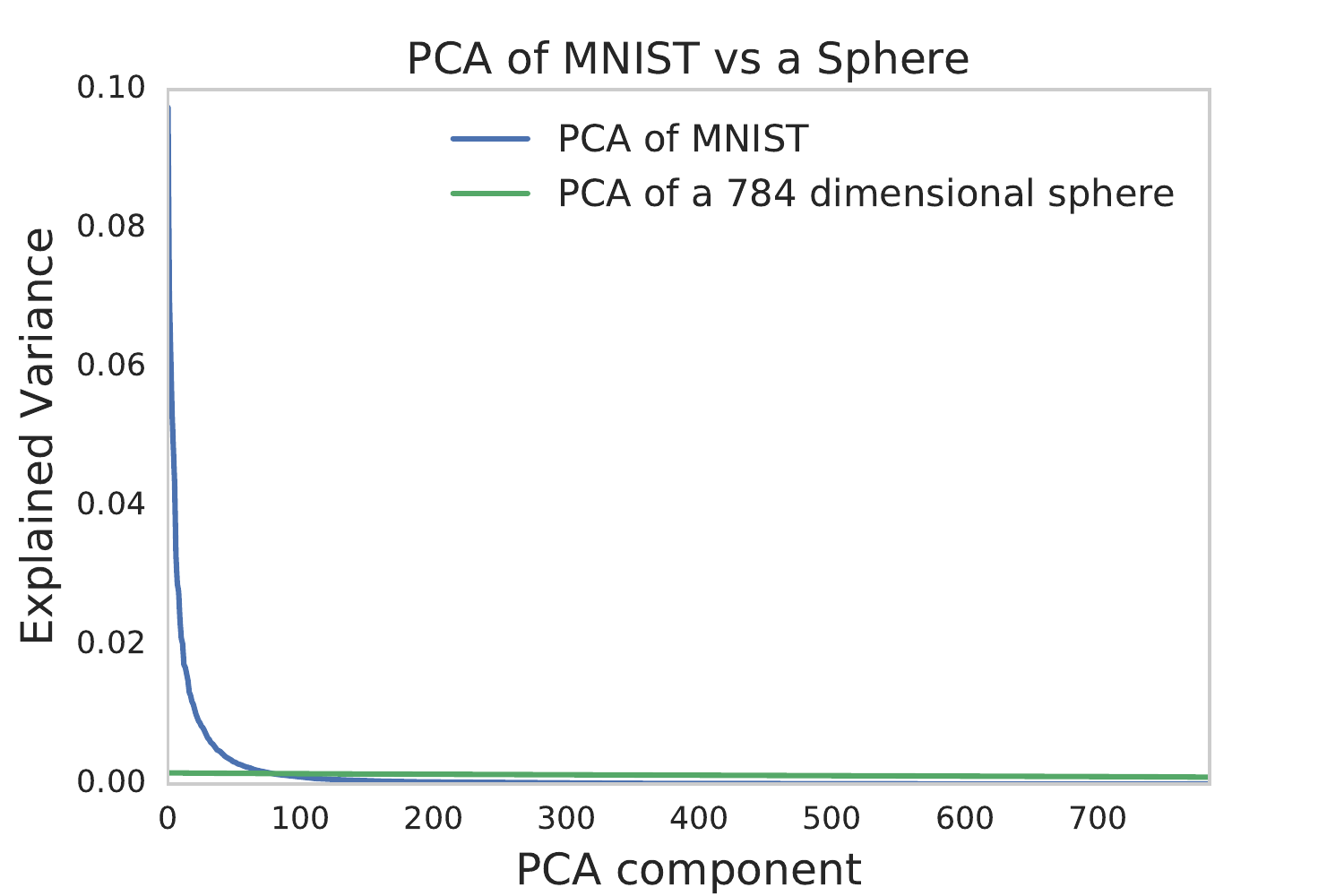}

\end{subfigure}%
\begin{subfigure}{.5\textwidth}
  \centering
  \includegraphics[width=1.0\linewidth]{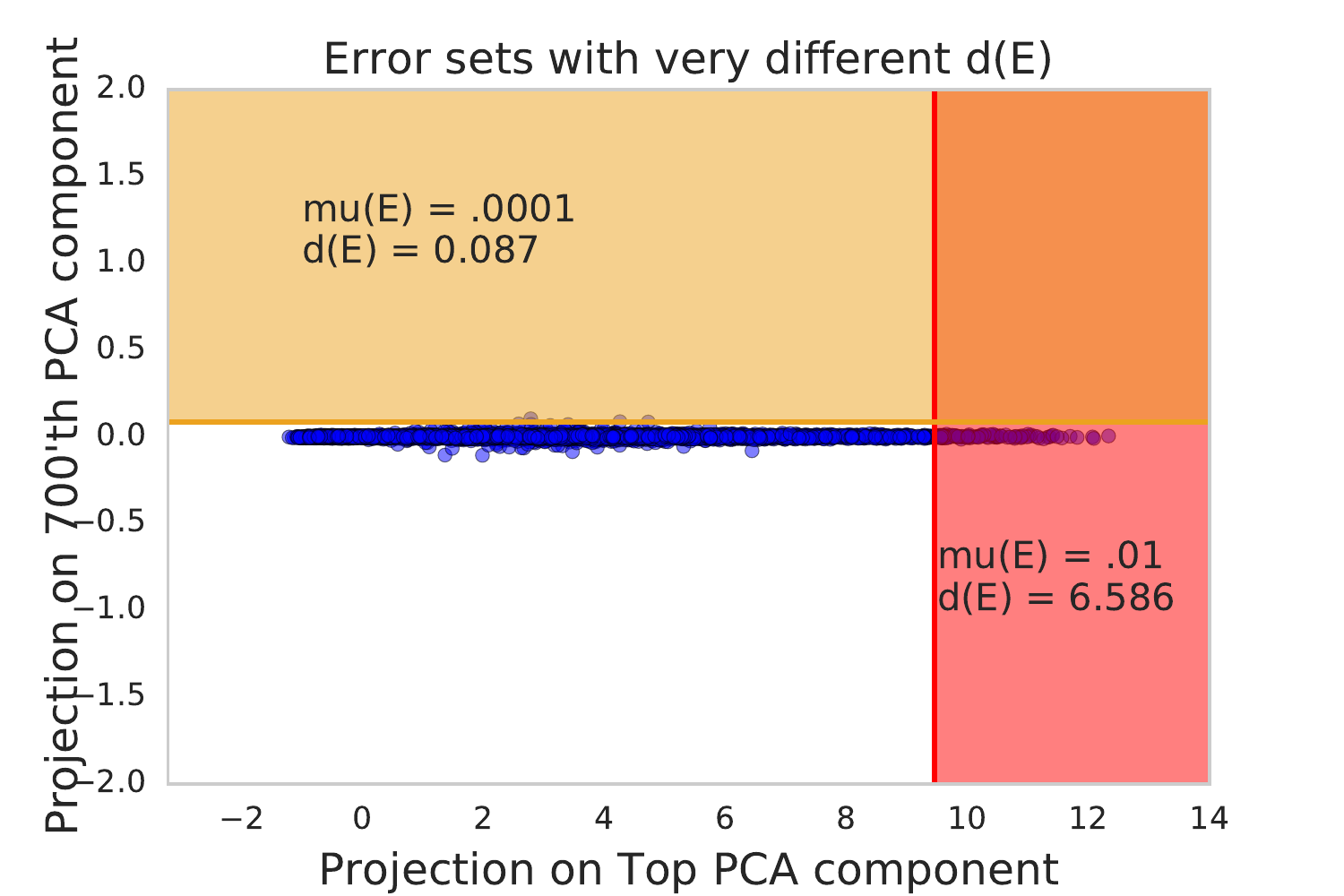}
\end{subfigure}

\caption{\textbf{Left:} To understand how Theorem 5.1 for the sphere relates to MNIST we compare the PCA explained variance ratios of the two datasets. The sphere is rotationally symmetric so all directions share the same explained variance. MNIST, however, is highly stretched along a few PCA directions. \textbf{Center:} We consider two different halfspace error sets which give vastly different relationships between $\mu(E)$ and $\de$. One way to maximize $d(E)$ given a fixed $\mu(E)$ is to distribute the errors in halfspaces aligned with the top PCA directions. The top PCA halfspace (red) has measure $\mu(E) = .01$ and $\de = 6.5$. The low variance PCA halfspace (orange) has small measure in the data distribution ($\mu(E) = .0001$) but also small average adversarial distance $d(E) = .084$.}
\label{fig:pca_mnist}
\end{figure*}

The goal of this section is to better understand whether any variant of Theorem 5.1 holds for real world datasets. This theorem is somewhat remarkable because it makes absolutely no assumption about the model decision boundary, other than there is test error $\mu(E)$. Thus it is simply a statement about the geometry of the data manifold. Rigorously proving a similar statement for real world datasets may be extremely difficult without a mathematical characterization of the data generating process $p(x)$. Nonetheless we can interpret the MNIST dataset as a finite iid sample from some theoretical distribution over images $p(x)$ and construct a set $E \subseteq \mathbb{R}^n$ which satisfies $\mu(E)$ is small but $d(E)$ is large. Note that simply choosing $E = \{x_1, \dots, x_k\}$ for 1\% of the test samples would not work here, because we should assume that the measure $\mu(\{x_i\})$ for any single image to be exceedingly small. Instead we inspect the training set and construct a set $E \subset \mathbb{R}^{784}$ for which 1\% of the training set lies in $E$, and for which $d(E)$ is large. Then we can interpret the test set as a fresh iid sample from $p(x)$ and estimate $\mu(E)$ and $d(E)$ from this sample. Using this approach we construct a set $E \subset \mathbb{R}^{784}$ for which $\mu(E) = .01$ and $d(E) = 6.59$, this set can be thought of as a lower bound on any variant of Theorem 5.1 one could hope to prove. For example, our construction shows that there is no theorem of the form ``every subset $E \subseteq \mathbb{R}^{784}$ with $\mu(E) = .01$ satisfies $d(E) < 6.59$''. However, one could imagine making \emph{some} assumptions about $E$, given that it needs to arise naturally from the decision boundary of a machine learning model, and showing a stronger bound. This would be interesting future work.

To construct such an $E$ we consider all possible halfspaces of $\mathbb{R}^{784}$ and choose the one which satisfies $\mu(E) = .01$ and maximizes $d(E)$. Considering halfspaces is natural, in particular the sets of maximal $d(E)$ for the sphere dataset are all halfspaces intersected with the data manifold. Recall that a halfspace of $\mathbb{R}^n$ is a set of the form $E_{\bm{w},b} = \{ \bm{x} \in \mathbb{R}^n : \bm{w}\cdot \bm{x} > b \}$, where the vector $\bm{w}$ is the \emph{normal} vector of a halfspace.  For any $\bm{x} \in \mathbb{R}^n$ the distance from $\bm{x}$ to the nearest point in this space, $d(x, E_{\bm{w},b})$, is easily computed as $\max(b - \bm{w} \cdot \bm{x}, 0)/||\bm{w}||$. 

In order to choose the optimal halfspace $E$, we first perform a PCA analysis of the MNIST dataset and compare this to PCA of the sphere (Figure~\ref{fig:pca_mnist}). The sphere is rotationally symmetric, so all PCA coefficients are the same. This means that all halfspaces $E$ of a fixed $\mu(E)$ will share the same $d(E)$. The explained variance of MNIST, however, is dominated by the first 100 PCA directions. This means that most halfspaces $E$ would have very small $\mu(E)$ and small $d(E)$. To choose the optimal $E$ then, we simply pick $\bm{w}$ to be the top PCA coefficient, then choose $b$ such that $1\%$ of the training set is on one side of the halfspace $E_{\bm{w},b} = \{ \bm{x} \in \mathbb{R}^n : \bm{w}\cdot \bm{x} > b \}$. We then use the test set to estimate $\mu(E)$ and $d(E)$ for this set, and observe that $\mu(E) = .01$ and $d(E) = 6.59$. We visualize this set in Figure~\ref{fig:pca_mnist} and compare it to a halfspace which is aligned with a low variance PCA direction. 

The existence of this set shows that there is room to improve adversarial robustness without reducing test error. For example, a standard convolutional network we trained satisfies $\mu(E) = .007$ and $d(E) = 1.3$. The network reported in \cite{madry2017advexamples} empirically estimated $\mu(E) = .014$ and $d(E) = 5.0$ ($d(E)$ here is in the $l_2$ metric). Note that calculating the true $d(E)$ is quite difficult, so we expect the true $d(E)$ of this model to be somewhat smaller. Nonetheless, external attempts to falsify the claims made in this paper have so far failed, so it is reasonable to believe that this model did in fact increase $d(E)$ somewhat. Errors still exist, however, local to most data points \cite{chen2017madry}.

\end{document}